\documentclass[a4paper,twoside]{article}

\usepackage{epsfig}
\usepackage[table]{xcolor}
\usepackage{subcaption}
\usepackage{calc}
\usepackage{amssymb}
\usepackage{amstext}
\usepackage{amsmath}
\usepackage{amsthm}
\usepackage{multicol}
\usepackage{pslatex}
\usepackage{apalike}
\usepackage{SCITEPRESS}     

\begin{document}

\title{Deep Learning-based Anomaly Detection on X-ray Images of Fuel Cell Electrodes}

\author{\authorname{Simon B. Jensen\sup{1}, 
Thomas B. Moeslund\sup{1}\orcidAuthor{0000-0001-7584-5209} and 
Søren J. Andreasen\sup{2}}
\affiliation{\sup{1}Department of Architecture and Media Technology, Aalborg University, Aalborg, Denmark}
\affiliation{\sup{2}Serenergy, Aalborg, Denmark}
\email{\{sbje,tbm\}@create.aau.dk, sja@serenergy.com}
}



\abstract{Anomaly detection in X-ray images has been an active and lasting research area in the last decades, especially in the domain of medical X-ray images. For this work, we created a real-world labeled anomaly dataset, consisting of 16-bit X-ray image data of fuel cell electrodes coated with a platinum catalyst solution and perform anomaly detection on the dataset using a deep learning approach. The dataset contains a diverse set of anomalies with 11 identified common anomalies where the electrodes contain e.g. scratches, bubbles, smudges etc. We experiment with 16-bit image to 8-bit image conversion methods to utilize pre-trained Convolutional Neural Networks as feature extractors (transfer learning) and find that we achieve the best performance by maximizing the contrasts globally across the dataset during the 16-bit to 8-bit conversion, through histogram equalization. We group the fuel cell electrodes with anomalies into a single class called abnormal and the normal fuel cell electrodes into a class called normal, thereby abstracting the anomaly detection problem into a binary classification problem. We achieve a balanced accuracy of 85.18\%. The anomaly detection is used by the company, Serenergy, for optimizing the time spend on the quality control of the fuel cell electrodes.} 

\onecolumn \maketitle \normalsize \setcounter{footnote}{0} \vfill

\section{\uppercase{Introduction}}
\label{sec:introduction}

\begin{figure}[h]
  \centering
  \includegraphics[width=0.4\textwidth]{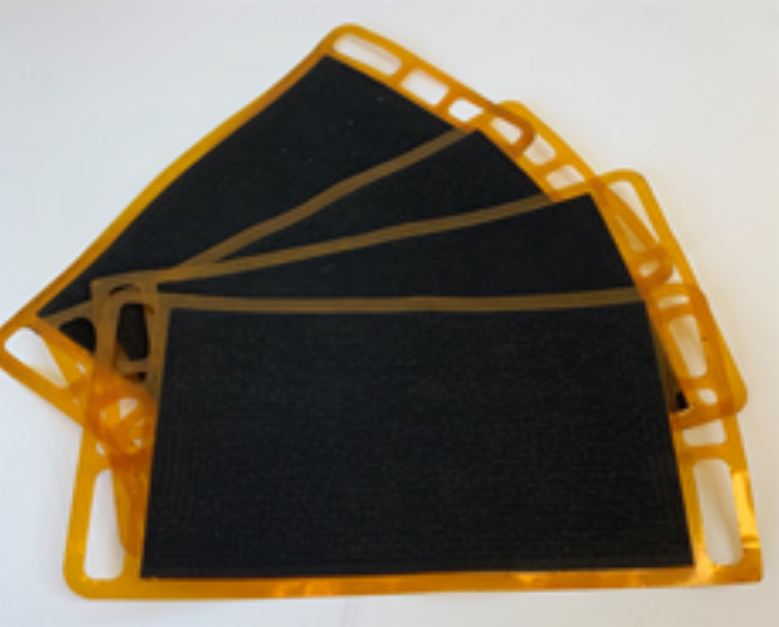}
  \caption{Fuel cell electrodes coated with a platinum catalyst}
  \label{fig:electrode-examples}
\end{figure}

Serenergy is a world-leading supplier of methanol-based fuel cell solutions with more than a thousand active units deployed globally. 
The fuel cells provide back-up power as well as temporary primary power or work in a hybrid system with renewable sources such as solar and/or wind. 
The core component of the fuel cell system, Serenergy, is a cell stack of 120 high temperature, polymer electrolyte membranes. Each cell contains 2 fuel cell electrodes that are coated with a platinum based catalyst.
Meaning the fuel cell system is made up of $120 \times 2 = 240$ fuel cell electrodes in total.
Examples of fuel cell electrodes can be seen in figure \ref{fig:electrode-examples}.
We will use the term fuel cell electrode or electrode interchangeably. 

The quality of the platinum based catalyst and the quality of how well it is coated on to each electrode is paramount to the overall conductivity of the fuel cell.
The quality of an electrode is measured through a semi-automatic/manual process where an X-ray image is manually captured of each electrode and the X-ray image is analyzed by an image analysis tool which outputs several quality parameters e.g., color histograms, box plots, standard deviation of the colors of the platinum coating, the minimum- and maximum colors of the platinum coating etc.

The process of analyzing the output of the image analysis tool is time consuming and in most cases the electrodes have an acceptable quality. Serenergy wishes to optimize the quality control process by using a deep learning approach to perform automatic anomaly detection on the fuel cell electrodes, by grouping them into two classes, normal or abnormal. 
Where a normal electrode can be used in the final fuel cell system and an abnormal electrode cannot. 

\begin{figure*}
  \centering
   \begin{subfigure}[b]{0.18\textwidth}
         \centering
         \includegraphics[width=\textwidth]{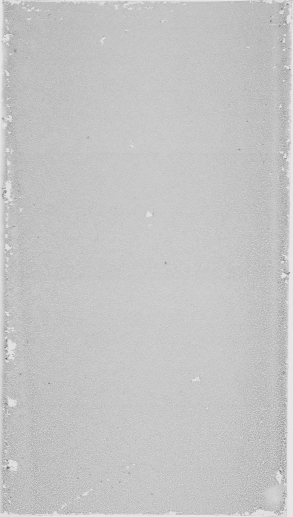}
         \caption{}
         \label{fig:x-ray-image-examples-a}
     \end{subfigure}
     \hfill
     \begin{subfigure}[b]{0.182\textwidth}
         \centering
         \includegraphics[width=\textwidth]{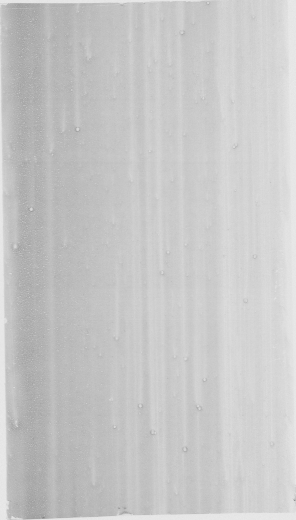}
         \caption{}         
         \label{fig:x-ray-image-examples-b}         
     \end{subfigure}
     \hfill
     \begin{subfigure}[b]{0.18\textwidth}
         \centering
         \includegraphics[width=\textwidth]{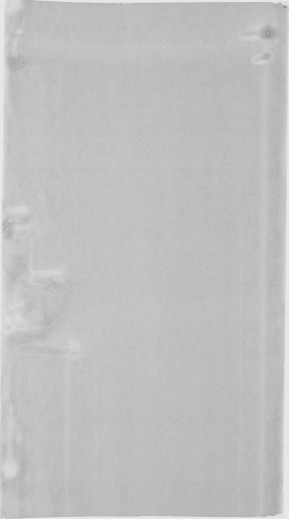}
         \caption{}
         \label{fig:x-ray-image-examples-c}
     \end{subfigure}
     \hfill
     \begin{subfigure}[b]{0.18\textwidth}
         \centering
         \includegraphics[width=\textwidth]{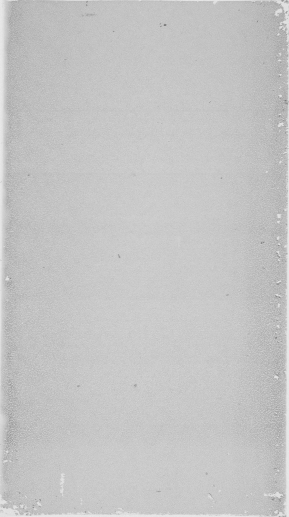}
         \caption{}
         \label{fig:x-ray-image-examples-d}
     \end{subfigure}
     \hfill
     \begin{subfigure}[b]{0.176\textwidth}
         \centering
         \includegraphics[width=\textwidth]{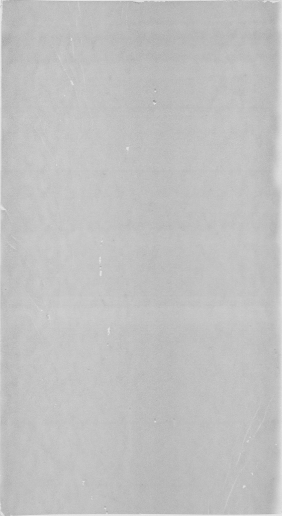}
         \caption{}
         \label{fig:x-ray-image-examples-e}
     \end{subfigure}
  \caption{Examples of X-ray images of fuel cell electrodes. Figure \ref{fig:x-ray-image-examples-a}, \ref{fig:x-ray-image-examples-b} and \ref{fig:x-ray-image-examples-c} are examples of abnormal plates, while figure \ref{fig:x-ray-image-examples-d} and \ref{fig:x-ray-image-examples-e} are examples of normal plates. \ref{fig:x-ray-image-examples-a} is abnormal due to issues with scratches near its edges. \ref{fig:x-ray-image-examples-b} is abnormal due to issues with lines and bubbles and \ref{fig:x-ray-image-examples-b} is abnormal due to issues with smudges.}
  \label{fig:x-ray-image-examples}
 \end{figure*}

In figure \ref{fig:x-ray-image-examples} examples of X-ray images of normal and abnormal electrodes are shown. This paper describes our approach to this important problem and the contributions are:

\begin{itemize}
    \item 16-bit to 8-bit conversion methods for X-ray images of fuel cell electrodes using histogram equalization.
    \item A Deep Convolutional Neural Network classifier to perform anomaly detection of X-ray images of fuel cell electrodes.
\end{itemize}

\subsection{Related Work}
\label{subsec:related-work}

\subsubsection{Deep Learning-based Classification on X-ray Image}
Deep learning is a widely used technology for classification of images. More specifically Deep Convolutional Neural Networks (CNNs). 
The technological development of CNNs has been accelerated by image classification datasets and challenges such as pascal VOC \cite{pascal-voc} and ImageNet \cite{imagenet}. 

For X-ray images, there has been a lack of large publicly available dataset, which means that few studies exists using CNNs for classification in X-ray images. 
However, in the specific domain of pneumonia detection in chest X-ray images, a number of dataset has recently been made publicly available for example \cite{chest-x-ray}, \cite{chest-x-ray-8} and \cite{covid-19-dataset}.  
This has resulted in a large number of studies which use CNNs to classify X-ray images in this specific field. \cite{pneumia-review} reviews and compares a large collection of these studies. 

\cite{pneumia-transfer-learning} trains four well-recognized CNN models pre-trained on the Imagenet dataset, AlexNet \cite{alexnet}, ResNet-18 \cite{resnet}, DenseNet201 \cite{densenet} and SqueezeNet \cite{squeezenet} for detecting pneumonia in the Pneumonia Chest X-ray dataset and compare their performance. They find that DenseNet201 achieves the best performance for binary classification (normal/pneumonia) while only performing slightly better than ResNet-18.

Similarly, \cite{pneumia-cnn} trains a CNN on the Pneumonia Chest X-ray dataset to detect pneumonia while utilizing the Dynamic Histogram Enhancement algorithm \cite{DHE} as pre-processing method to improve the quality of X-ray images before training and evaluating the CNN model.   

\subsubsection{Transfer Learning for Image Classification}
\label{subsubsec:transfer-learning}
\cite{transfer-learning-study} shows that transfer learning for image classification using deep CNNs is a valid and efficient method for achieving high performance in image classification tasks when dealing with datasets of limited size. They use the Inception-v3 \cite{inception-v3} CNN pre-trained on the ImageNet dataset and re-train the model on the Caltech Face dataset consisting of only 450 images while achieving an accuracy of 65.7\%. 

\cite{pneumia-transfer-learning} validates transfer learning for image classification when the base dataset (ImageNet) consist of 3-channel RGB images and the target dataset consists of 1-channel grayscale X-Ray images. They do this by fine-tuning CNN models pre-trained with ImageNet on chest X-ray datasets.  

Several transfer learning methods have recently been publicized which further optimizing the performance which can be achieved by the method. 
\cite{transfer-learning-pay-attention} proposes a method called attentive feature distillation and selection (AFDS),
which adjusts the strength of transfer learning regularization and also dynamically determines the important features to transfer. They impose the method onto ResNet-101 and achieve state-of-the art computation reduction. 

\subsubsection{Anomaly Detection using Deep Learning}
\cite{anomaly-detection-challenges} reviews twenty studies which utilizes deep learning for anomaly detection and identify challenges and insights in the domain. They identify the three main challenges for anomaly detection to be: (1) handling the class imbalance of normal and abnormal data, (2) the availability of labeled data and (3) the fact that there is often noise in the data that appears to be close to the actual anomalies and thus it becomes difficult to differentiate them.

\cite{weighted-cross-entropy} proposes a solution to the class imbalance challenge, by introducing a new loss function for binary- and multiclass classification problems called Real-World-Weight Cross-Entropy loss function. Which allows direct input of real world costs as weights. This could prove useful for classification problems where there is a well-defined loss/cost for misclassified samples. 

\cite{covid-19-data-augmentation} proposes a solution to the second challenge by introducing and comparing four data augmentation methods for artificially increasing the number of training samples of X-ray images, while performing Covid-19 pneumonia detection using a CNN. The methods uses combinations of random rotations, shear, translation, horizontal- and vertical flipping among other data augmentation methods. 

We discuss the three challenges and possible solutions further, in regards to anomaly detection in X-ray images of fuel cell electrodes in section \ref{sec:discussion}.

\section{\uppercase{Approach}}
\label{sec:approach}

\subsection{Overview}
\label{subsec:overview}
 
\begin{figure*}[!h]
  \centering
  \includegraphics[width=\textwidth]{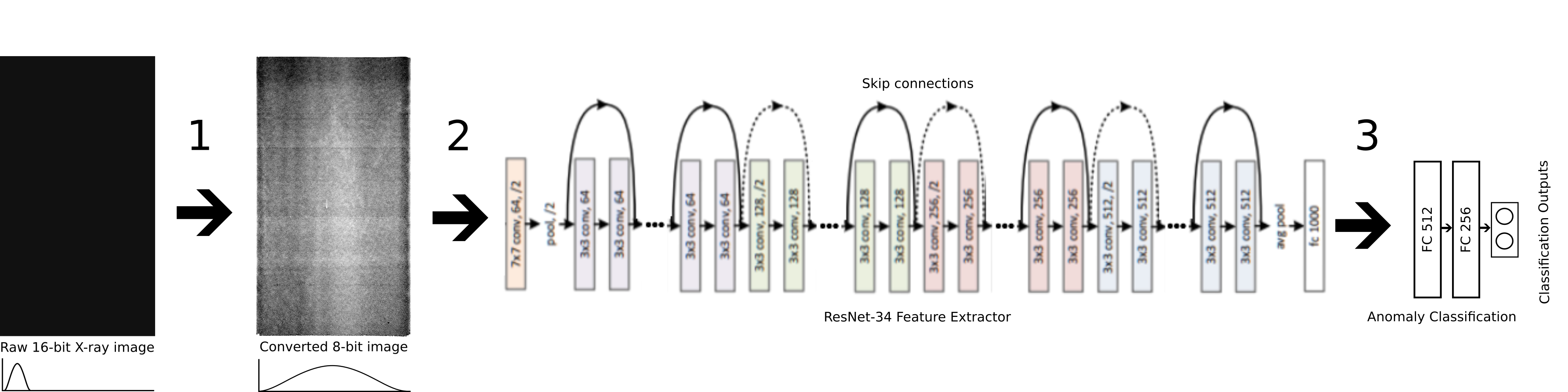}
  \caption{An overview of the anomaly detection approach proposed in this paper made up by three main steps which are described in section \ref{sec:approach}. Some of the layers of the ResNet-34 feature extractor are hidden for illustration purposes.}
  \label{fig:approach-overview}
 \end{figure*}

An overview of the anomaly detection approach proposed in this paper is seen in figure \ref{fig:approach-overview}. The approach is described by the following three steps:

\begin{enumerate}
\item Convert the X-ray images from 16-bit to 8-bit, as described in section \ref{subsec:16-bit-to-8-bit-conversion}.
\item Extract features using a pre-trained ResNet-34 CNN model as described in \ref{subsec:feature-extraction-and-detections}.
\item Classify anomalies in the feature-maps generated by the CNN model, using fully connected layers, as described in \ref{subsubsec:training}. 
\end{enumerate}

\subsection{The Fuel Cell Electrode X-Ray Dataset}
\label{subsec:electrode-plate-x-ray-data-set}
A real-world labeled anomaly dataset consisting of 16-bit X-ray images of platinum catalyst coated fuel cell electrodes is created for this work, as no existing dataset in the domain of fuel cell electrodes for methanol fuel cells exists to the best of our knowledge.

\begin{figure}[!h]
  \centering
  \includegraphics[width=0.45\textwidth]{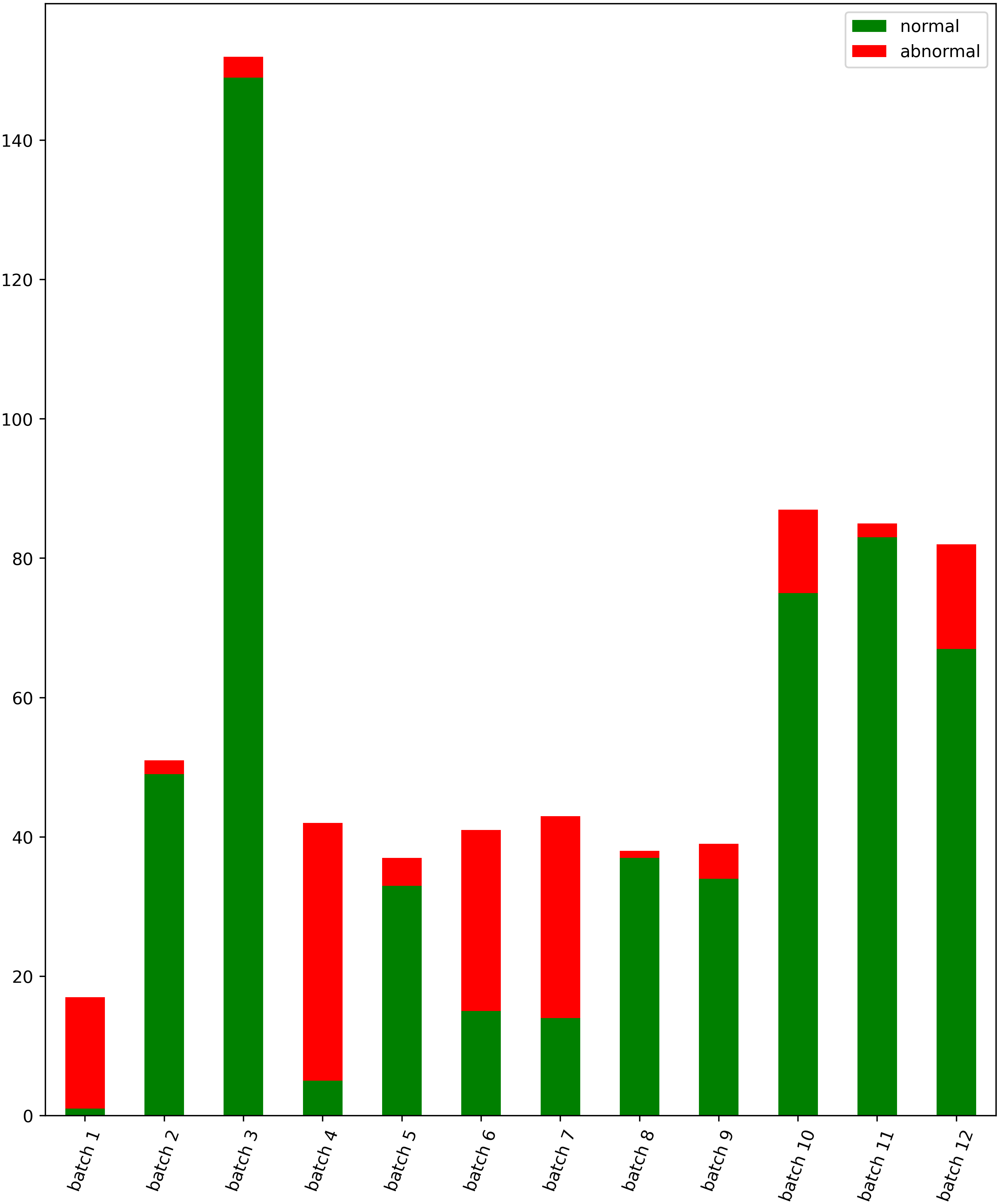}
  \caption{Overview of the number of X-ray images per batch as well as the number of normal and abnormal samples. The number of X-ray images per batch vary for each batch and so does the class balance. The dataset is generally imbalanced with a over-representation of normal samples.}
  \label{fig:batch-overview}
 \end{figure}

The fuel cell electrode X-ray dataset consists of 714 X-ray images. Each electrode belongs to 1 of 12 batches, named batch 1, batch 2, ..., batch 12. Where a batch represents a collection of electrodes which are coated with a platinum catalyst solution using the same coating method and mixture. The batches in the dataset have varying number of X-ray images as can be seen from figure \ref{fig:batch-overview}. Batch 3 has most sample with 152 and batch 1 has fewest samples with 17. The class balance vary greatly for each batch as well. For batch 1 only 1 electrode is normal accounting for ~5.88\% of the images in the batch while 37 out of the 38 electrodes in batch 8 are labeled as normal accounting for 97.37\% of the images in the batch. The dataset is generally imbalanced with a over representation of normal samples. Across all 714 images 562 (78.71\%) are labeled as normal and 152 (21.29\%) are labeled as abnormal.

A dataset size of 714 X-ray images is considered to be a small dataset when utilizing a deep learning approach. We utilize transfer-learning to overcome this problem, through the use of a pre-trained ResNet-34 model, as described in section \ref{subsec:feature-extraction-and-detections}. Due to the limited size of the dataset, creating a representative test set for the dataset proves difficult. We evaluated the X-ray images of each batch through cross-validation. This is done by evaluating each batch individually, while the remaining batches are used as training samples and combining the results of each evaluation into a total score. This is further described in section \ref{sec:results}. 

\subsubsection{Fuel Cell Electrode X-Ray Images}
\label{subsubsec:electrode-plate-x-ray-images}

The X-ray images of the dataset have varying dimensions with a minimum width of 1119 pixels, maximum width of 1219 pixels and mean width of 1145 pixels. The minimum height of an electrode is 2053 pixels, the maximum height is 2115 pixels and the mean 2072,1 pixels. During training and evaluation of the anomaly detector, the X-ray images are transformed into a uniform size of $2000\times1000$. Examples of normal and abnormal electrode X-ray images can be seen in figure \ref{fig:x-ray-image-examples}.

\subsubsection{Anomalies}
\label{subsubsec:anomalies}

Serenergy has identified 11 common anomaly types, which are grouped into a single class called abnormal. The identified 11 common anomaly types are named: scratches, lines, edge cuts, edge tensions, smudges, edge ink flow, bubbles, missing ink, agglomerate, ink fluctuations, ink entry/exit. 

The normal fuel cell electrodes can have minor representations of one or more anomaly type, as long as the severity is not too great. Figure \ref{fig:x-ray-image-examples-d} is an example of an electrode which have minor scratches, but the scratches are not sever enough to be classified as abnormal, while figure \ref{fig:x-ray-image-examples-a} is an example of a fuel cell electrode which is abnormal due to scratches near its edges.

\subsection{X-ray Image Conversion Methods}
\label{subsec:16-bit-to-8-bit-conversion}

To utilize pre-trained CNNs such as ResNet-34 as feature extractor in the anomaly detector, the depth of the electrode X-ray images is extended from 1 channel to 3 channels and the pixel values of the images are converted from a 16-bit values into 8-bit values. Thous, increasing the similarity of the electrode X-ray images to the images of ImageNet on which the CNN is pre-trained on. This is further described in section \ref{subsec:feature-extraction-and-detections}.

The 16-bit color range consists of 2 to the power of 16 (65536) colors and the 8-bit color range consists of 2 to the power of 8 (256) colors. A loss of information during the conversion is therefore inevitable. 

We experiment with four methods for converting the X-ray images from 16- to 8-bits using histogram equalization and implemented using Python's OpenCV library \cite{opencv}.

\begin{figure*}
  \centering
   \begin{subfigure}[b]{0.22\textwidth}
         \centering
         \includegraphics[width=\textwidth]{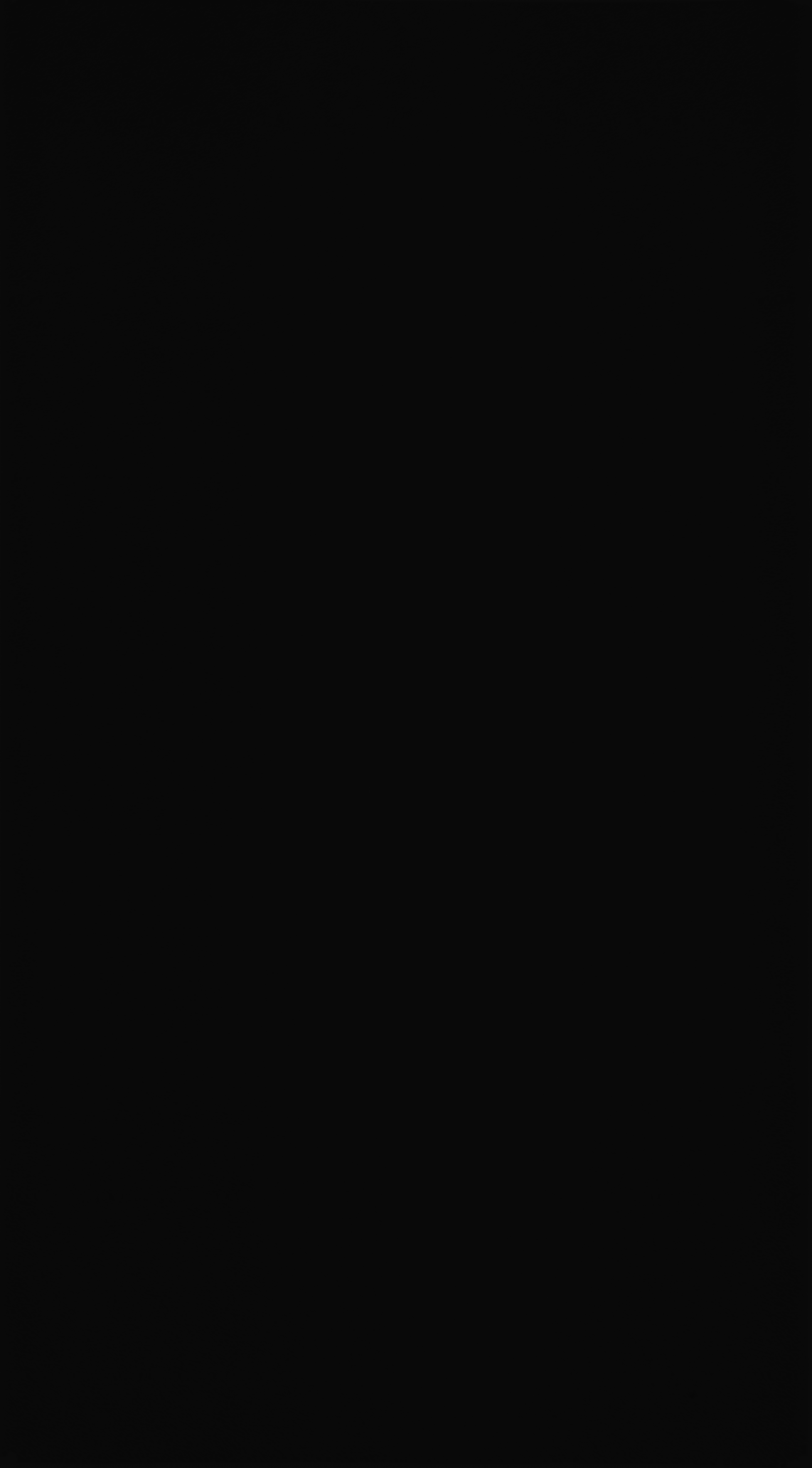}
         \caption{}
         \label{fig:preprocessed-x-ray-image-example-a}
     \end{subfigure}
     \hfill
     \begin{subfigure}[b]{0.22\textwidth}
         \centering
         \includegraphics[width=\textwidth]{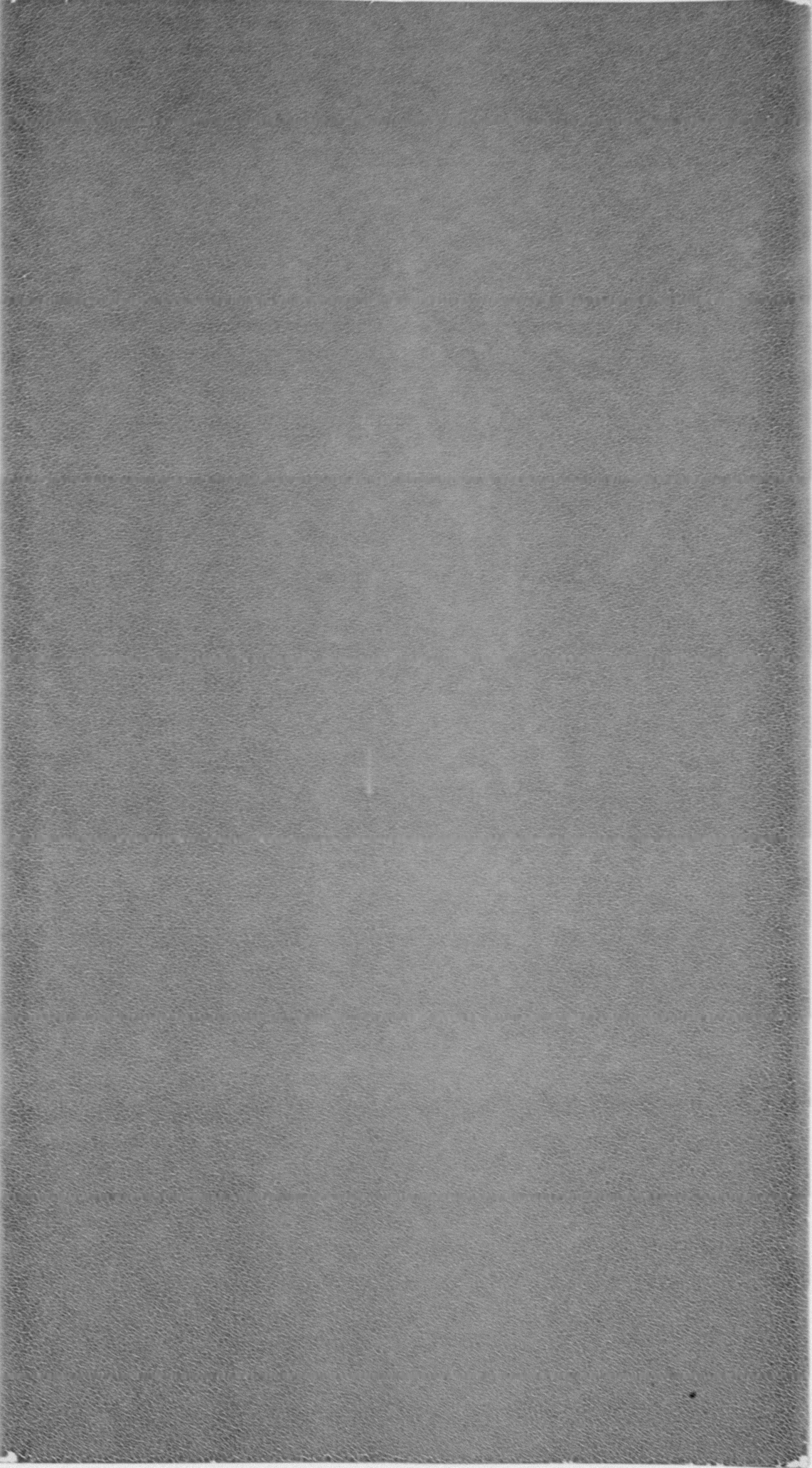}
         \caption{}
         \label{fig:preprocessed-x-ray-image-example-b}
     \end{subfigure}
     \hfill
     \begin{subfigure}[b]{0.22\textwidth}
         \centering
         \includegraphics[width=\textwidth]{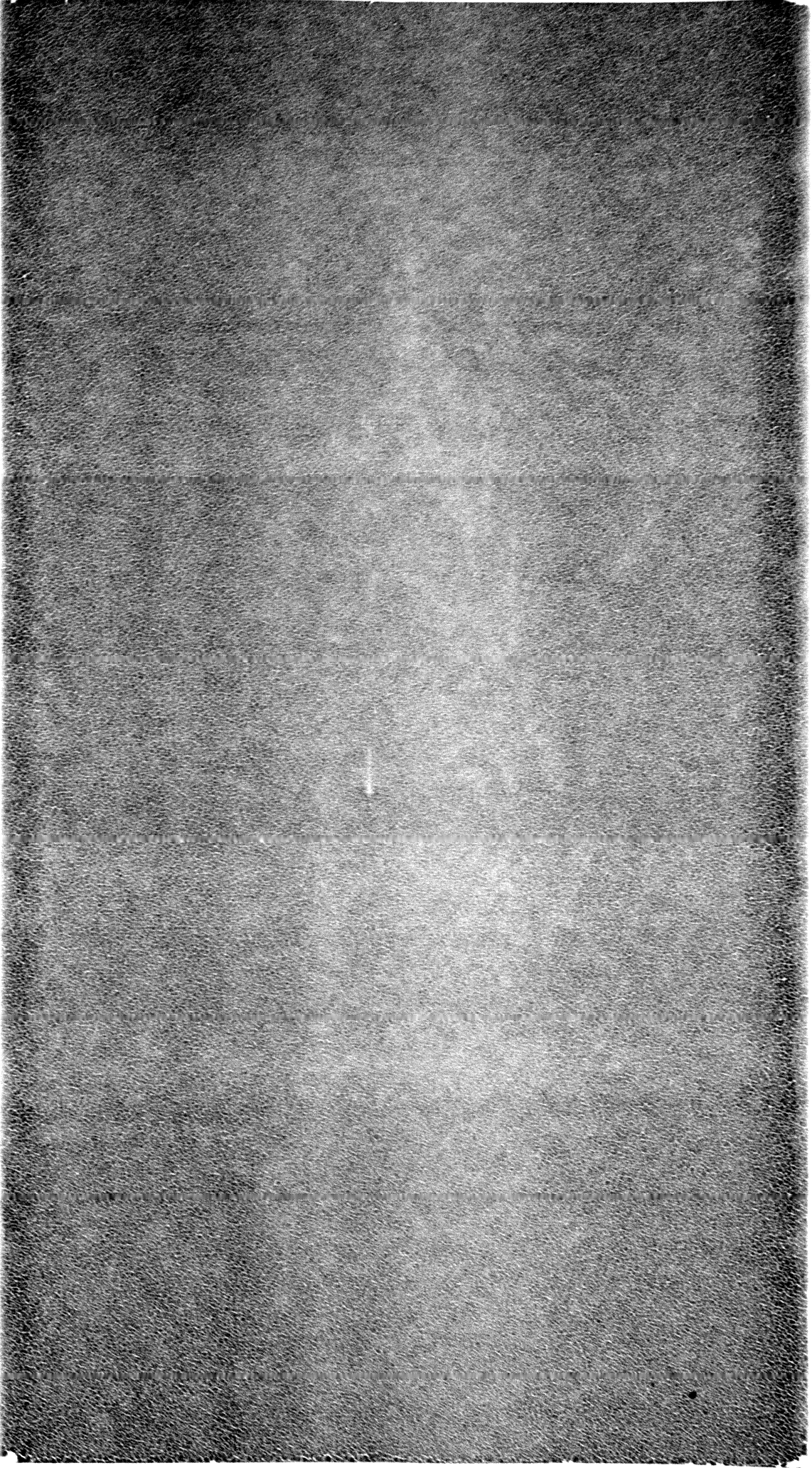}
         \caption{}
         \label{fig:preprocessed-x-ray-image-example-c}
     \end{subfigure}
     \hfill
     \begin{subfigure}[b]{0.22\textwidth}
         \centering
         \includegraphics[width=\textwidth]{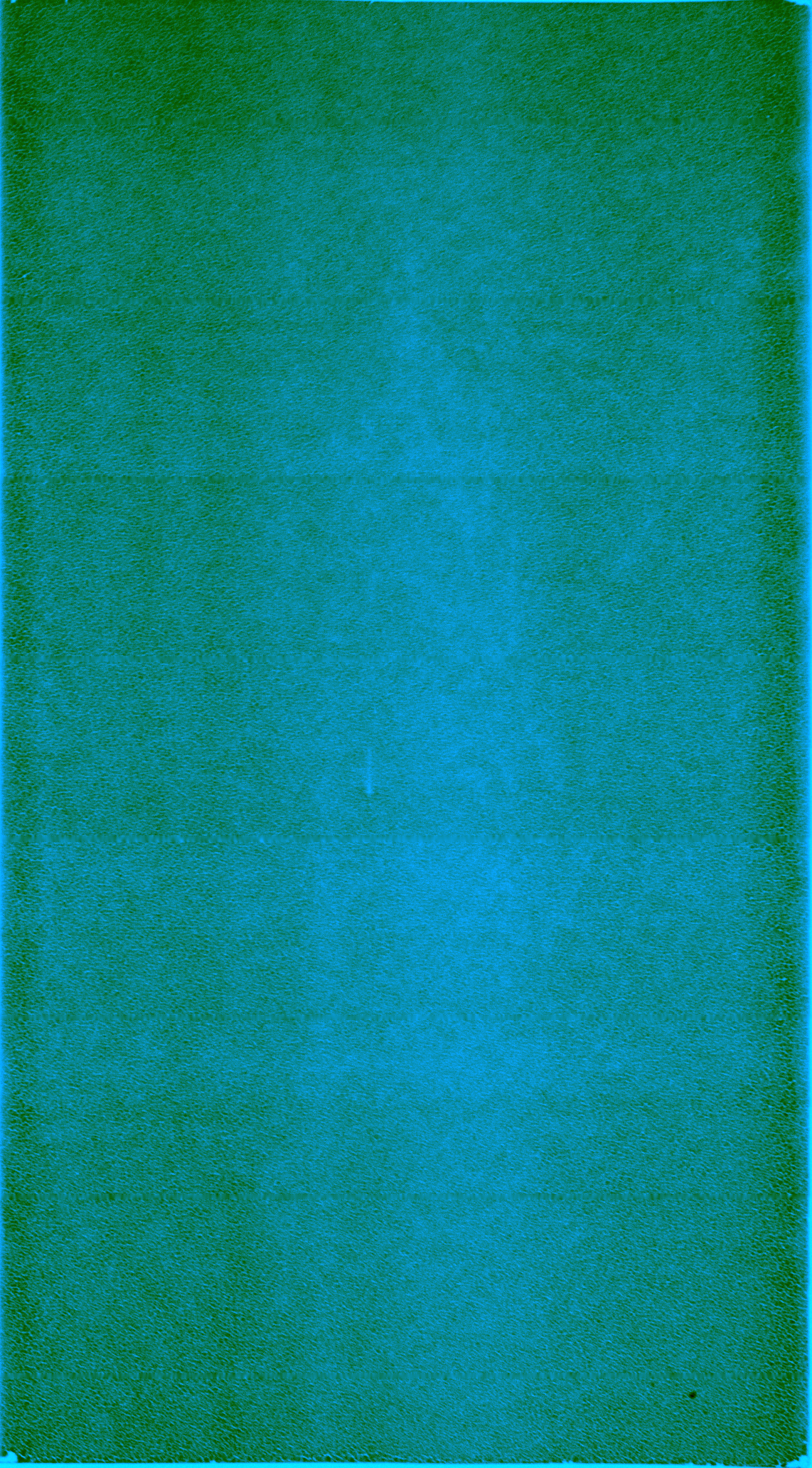}
         \caption{}
         \label{fig:preprocessed-x-ray-image-example-d}
     \end{subfigure}
  \caption{Example of the 8-bit images obtained by applying conversion method 1, 2, 3 and 4 to a 16-bit X-ray image. Figure \ref{fig:preprocessed-x-ray-image-example-a}, \ref{fig:preprocessed-x-ray-image-example-b}, \ref{fig:preprocessed-x-ray-image-example-c} and \ref{fig:preprocessed-x-ray-image-example-d} shows method 1, 2, 3 and 4 respectively.}
  \label{fig:preprocessed-x-ray-image-examples}
 \end{figure*}

In section \ref{subsec:method-1-naive-conversion} a naive 16-bit to 8-bit conversion method is described  which is used as baseline.
In section \ref{subsec:method-2-conversion-by-global-min-and-max} we use histogram equalization with global maximum and minimum bounds calculated across the entire dataset and in section \ref{subsec:method-3-conversion-by-local-min-and-max} we use histogram equalization with local maximum and minimum bounds calculated for each individual X-ray image. 
Finally, we mix the methods in section \ref{subsec:method-4-mixing-the-methods}. 
Examples of the resulting 8-bit images for each conversion are shown in figure \ref{fig:preprocessed-x-ray-image-examples}. 

\subsubsection{Method 1: Naive Conversion}
\label{subsec:method-1-naive-conversion}
Method 1 is a naive conversion which is used as baseline. Method 1 simply scales each 16-bit pixel value and convert it to an unsigned 8-bit type. The conversion is given by equation \ref{eq:method-1}.

\begin{equation}\label{eq:method-1}
    f_{1}(x_{i})=\text{usign}\left(\frac{x_{i}}{256}\right)
\end{equation}

Where $x_{i}$ corresponds to the $i$th pixel in a 16-bit X-ray image and $\text{usign}$ is a function which converts a number to an 8-bit unsigned integer. An example of a resulting fuel cell electrode image after conversion can be seen in figure \ref{fig:preprocessed-x-ray-image-example-a}.

Most pixels lie in the range 1700-2800 in the 16-bit X-ray images, as shown in section \ref{subsec:method-2-conversion-by-global-min-and-max}, which means method 1 will appear very dark. Pixel values in the 16-bit color range of 1700-2800 corresponds to pixel values of ~6-11 in the 8-bit range using naive conversion.

Finally, the resulting 1-channel 8-bit image is extended with 2 additional channel, resulting in a 3. channel 8-bit image. 

\subsubsection{Method 2: Conversion by global min and max}
\label{subsec:method-2-conversion-by-global-min-and-max}
For method 2 the global maximum- and minimum pixel value, $G_{max}$ and $G_{min}$, of the 16-bit X-ray dataset is calculated and used as upper- and lower-bounds for histogram equalization, during 16-bit to 8-bit conversion. 

The global maximum pixel value is found by calculating the 99.99th percentile of the pixel values in all X-ray images in the dataset and taking the maximum pixel value found and round it to nearest hundred. 
Similarly, we calculate $G_{min}$ by finding the 0.01th percentile. $G_{max}$ and $G_{min}$ are found to be 28000 and 1700. Method 2 is given by equation \ref{eq:method-2} and illustrated in figure \ref{fig:histogram-min-max}. 

\begin{equation}\label{eq:method-2}
    f_{2}(x_{i})=\text{usign}\left(\frac{\max(\min(x_{i},G_{max}),G_{min})}{G_{max}-G_{min}} \times 256\right)
\end{equation}

The reason for calculating the 99.99th and 0.01th percentiles of the pixel values is to avoid that noise in the X-ray images will affect the values of $G_{max}$ and $G_{min}$.

\begin{figure}[!h]
  \centering
  \includegraphics[width=0.45\textwidth]{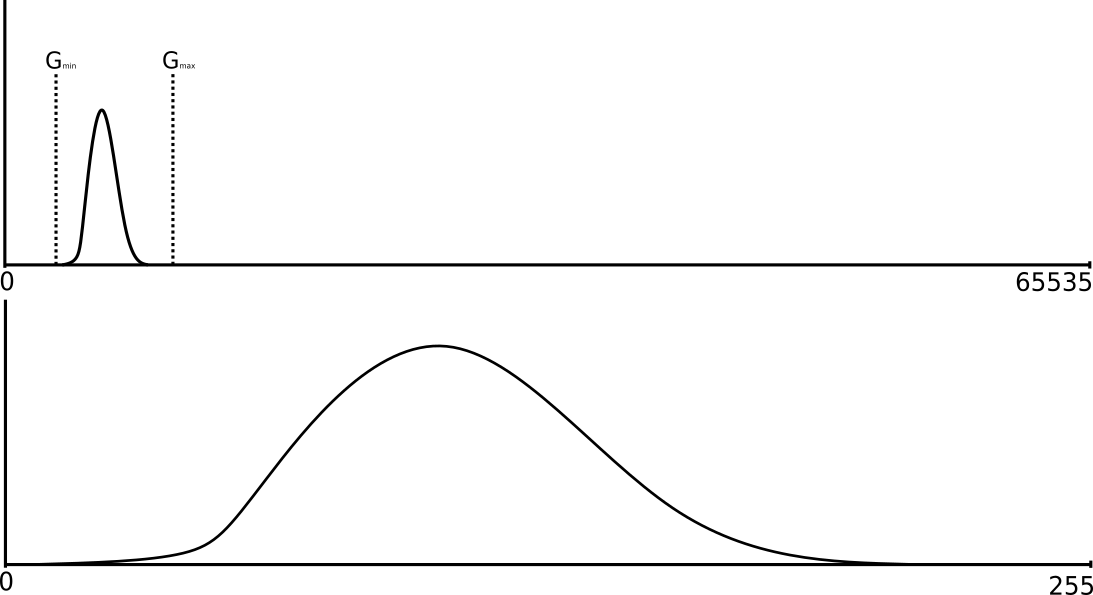}
  \caption{Method 2 uses the global minimum- and maximum pixel values, $G_{max}$ and $G_{min}$, calculated across the entire fuel cell electrode X-ray dataset as upper- and lower bounds to convert from 16-bit to 8-bit through histogram equalization.}
  \label{fig:histogram-min-max}
\end{figure}

\subsubsection{Method 3: Conversion by local min and max}
\label{subsec:method-3-conversion-by-local-min-and-max}
Method 3 uses the local maximum- and minimum pixel value, $L_{max}$ and $L_{min}$, found for each individual X-ray image, as upper- and lower-bound for histogram equalization. 

Thereby, maximizing the contrast in the 8-bit color range for each individual X-ray image. Similarly to method 2, the 99.99th and 0.01th percentiles of the pixel values for each X-ray image are used to avoid that noise will affect the values of $L_{max}$ and $L_{min}$.

The danger however, can be that the converted pixel values in one X-ray image lose there meaning relative to the converted pixel values of another X-ray image. 
Consider an X-ray image, $\text{image}_i$ with a $L_{min}$ value of 2200 and another X-ray image, $\text{image}_j$ with an $L_{min}$ value of 1800. After conversion, a value of 0 in the 8-bit color range will have corresponded to 2200 in $\text{image}_i$ and to 1800 in $\text{image}_j$. Method 3 is given by equation \ref{eq:method-3}. 

\begin{equation}\label{eq:method-3}
\begin{split}
\begin{aligned}
    f_{3}(k, x_{i}) = \text{usign}\left(\frac{\max(\min(x_{i},L_{k,max}),L_{k,min})}{L_{k,max}-L_{k,min}} \times 256\right)
\end{aligned}
\end{split}
\end{equation} 

Where $L_{k,max}$ and $L_{k,min}$ correspond to the local maximum and minimum of the k'th image in the electrode dataset.

\begin{figure}[!h]
  \centering
  \includegraphics[width=0.45\textwidth]{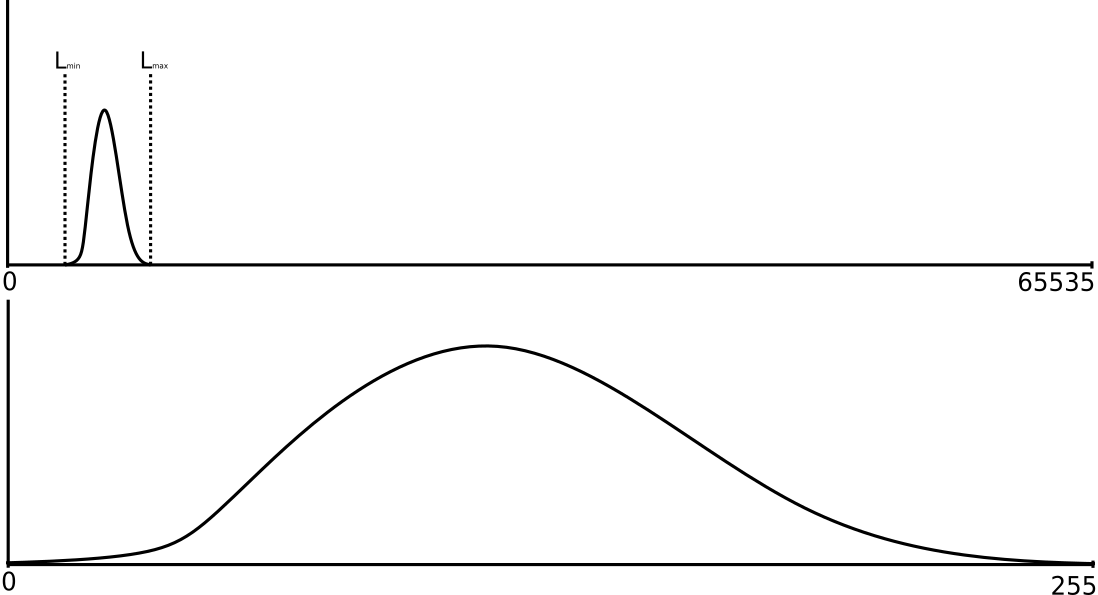}
  \caption{Method 3 uses the local minimum- and maximum pixel values, $L_{max}$ and $L_{min}$, calculated for each individual electrode X-ray image as upper- and lower bounds to convert from 16-bit to 8-bit through histogram equalization. Thereby, maximizing the contrast in the 8-bit color range for each individual X-ray image.}
  \label{fig:histogram-min-max}
\end{figure}

\subsubsection{Method 4: Mixing the methods}
\label{subsec:method-4-mixing-the-methods}
\begin{figure}[!h]
  \centering
   \includegraphics[width=0.25\textwidth]{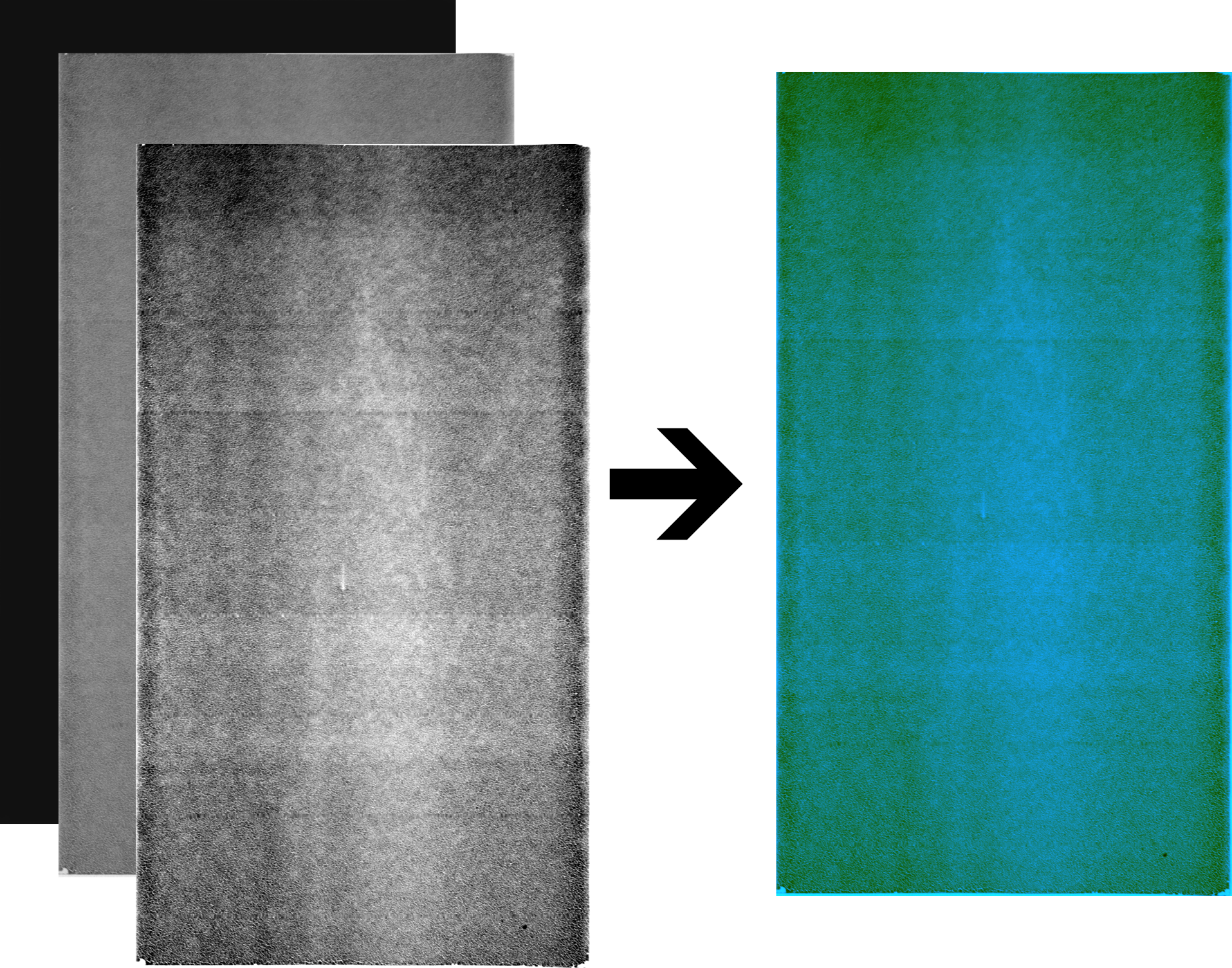}
  \caption{Method 4 mixes the conversion methods from method 1, 2, 3 into a 3 channel image, with 1 channel corresponding to the values achieved by each of the methods.}
  \label{fig:method-4-example}
 \end{figure}

Finally, method 4 mixes the conversion methods from method 1, 2 and 3, such that the resulting image will contain the 8-bit pixel values from method 1 in its 1. channel, the pixel values from method 2 in its 2. channel and the pixel values from method 3 in its 3. channel as seen in figure \ref{fig:method-4-example}.

\subsection{Anomaly Detector Architecture}
\label{subsec:feature-extraction-and-detections}
This paper proposes a fuel cell electrode anomaly detector which uses a Convolutional Neural Network as feature extractor. For this purpose, a PyTorch \cite{pytorch} implementation of the ResNet-34 \cite{resnet} CNN model is utilized.

The ResNet-34 model is pre-trained on a base dataset (ImageNet \cite{imagenet}).
The reason for using a pre-trained CNN, also referred to as transfer learning, is that training CNNs from random initializations, usually requires a large amount of data, to achieve a high performance. 
For many real-world applications it can be both time consuming and expensive to collect the required amount of data, as is also the case for this research.

The ResNet-34 model is extended with two fully connected layers of size 512 and 256 and finally with a softmax layer of size 2, to get an output for each class, normal and abnormal. The model architecture is illustrated in figure \ref{fig:approach-overview}.  

\subsubsection{Training}
\label{subsubsec:training}

The anomaly detector model is trained on a Nvidia GeForce GTX 1080 GPU for $12\times50$ epochs for each of the 4 conversion methods. All layers in the ResNet-34 CNN are frozen, accept for the fully connected layer, which is fine-tuned together with the fully connected layers of the anomaly classifier. We use the cross entropy loss as loss function and an initial learning rate of 0.001 which decays for each 10 epochs.  

We augment the dataset with random horizontal and vertical flips (set to 50\% probability) and resize the electrode images to height of 2000 pixels and width of 1000 pixels.

\section{\uppercase{Results}}
\label{sec:results}
In this section we present the results achieved by the anomaly detector when applying the 4 conversion methods described in section \ref{subsec:16-bit-to-8-bit-conversion} to the fuel cell electrode X-ray dataset. The anomaly detectors are evaluated using the balanced accuracy metric described in section \ref{subsec:balanced-accuracy}.

\subsection{Balanced accuracy}
\label{subsec:balanced-accuracy}
We use the balanced accuracy metric to evaluate the anomaly detector, as it proves useful for binary classification problems on datasets with class imbalance, like in the fuel cell electrode dataset. Whereas, accuracy can be misleading if the class imbalance is great. The balanced accuracy metric overcomes this issue by weighting the positive and negative samples equally significant despite one class being more numerous than the other. This is done by adding the true positive rate (TPR) with the true negative rate (TNR) and dividing them by 2, as can be seen in equation \ref{eq:balanced-accuracy}.


\begin{equation}\label{eq:balanced-accuracy}
    \text{BALANCED ACC}=\frac{TPR + TNR}{2}
\end{equation}

\begin{equation}\label{eq:true-positive-rate}
    \text{TPR}=\frac{TP}{TP + FN}
\end{equation}

\begin{equation}\label{eq:true-negative-rate}
    \text{TNR}=\frac{TN}{TN + FN}
\end{equation}

For a dataset with 98 positive samples and 2 negative samples, a classifier will achieve an accuracy score of 98\% by simply classifying every sample as positive. The balanced accuracy score will only be 50\% in this case.

\subsection{Cross-validation Evaluation}
\label{subsec:cross-validation-evaluation}
The dataset is evaluated through cross-validation, due to the limited size of the dataset and the nature of the dataset where each batch of images are coated with a platinum catalyst solution using the same coating method and solution mixture. This means images from the same batches will have a high similarity to one another. Utilizing samples from the same batch as both training and testing samples will therefore inevitable occlude the performance of the anomaly detectors. 

We train the anomaly detector as described in section \ref{subsubsec:training} using all but one batch, which is used as test set. The same procedure is replicated until each batch has been evaluated individually for each of the 4 conversion methods. Meaning a total of $4 \times 12$ training/evaluations are performed. For each training/evaluation run, the balanced accuracy score is calculated. A combined balanced accuracy score is then calculated for each conversion method by adding the true positives (TP), false positives (FP), true negatives (TN) and false negatives (FN) obtained when evaluating each of the 12 batches, for the given conversion method. The results can be seen in table \ref{tab:evaluation-results-balanced-acc}.

We find that the best anomaly detection performance is achieved by conversion method 2, with a balanced accuracy of 85.18\%. Surprisingly, we find that method 4, which combines conversion method 1, 2 and 3 achieves the worst overall anomaly detection performance. A possible explanation for this might be that the dissimilarity between the combined 3-channel image created by method 4 and the features of RGB images in ImageNet, which our feature extractor CNN is pre-trained on, is too great.  

\begin{table*}[h]
\caption{The balanced accuracy (\%) achieved by the anomaly detector when applying conversion method 1, 2, 3 and 4 to the fuel cell electrode X-ray images. The performance achieved by each conversion method is measured through cross-validation, where each batch is used as test set individually while the remaining batches are used as training set. The green colors highlight the method(s) which achieved the best balanced accuracy score for each batch. The olive green color highlights the method which achieved the best overall balanced accuracy score when combining the evaluations of each batch, which was conversion method 2.}
\label{tab:evaluation-results-balanced-acc} 
\centering
\resizebox{0.6\textwidth}{!}{
\begin{tabular}{|c|c|c|c|c|c|}
  \hline
  \textbf{Test Batch} & \textbf{Method 1} & \textbf{Method 2} & \textbf{Method 3} & \textbf{Method 4} \\
  \hline
  	Batch 1 		& 50.00 & \cellcolor{green!25} 96.88 & 50.00 & 40.63 \\
  \hline
    Batch 2 		& \cellcolor{green!25} 75.00 & 50.00 & \cellcolor{green!25} 75.00 & \cellcolor{green!25} 75.00 \\
  \hline
    Batch 3 		& \cellcolor{green!25} 50.00 & \cellcolor{green!25} 50.00 & \cellcolor{green!25} 50.00 & \cellcolor{green!25} 50.00 \\
  \hline
    Batch 4 		& \cellcolor{green!25} 90.54 & 58.65 & 67.30 & 67.30 \\
  \hline
    Batch 5 		& \cellcolor{green!25} 75.00 & 62.50 & \cellcolor{green!25} 75.00 & 62.50 \\
  \hline
    Batch 6 		& 66.15 & \cellcolor{green!25} 87.05 & 77.56 & 85.13 \\
  \hline
    Batch 7 		& 82.39 & \cellcolor{green!25} 89.53 & 73.89 & 72.17 \\
  \hline
    Batch 8 		& 50.00 & 50.00 & \cellcolor{green!25} 100.00 & 50.00 \\
  \hline
    Batch 9 		& 68.53 & 50.00 & \cellcolor{green!25} 90.00 & 70.00 \\
  \hline
    Batch 10 		& 50.00 & 50.00 & \cellcolor{green!25} 54.17 & 54.17 \\
  \hline
   Batch 11			& 75.00 & \cellcolor{green!25} 99.40 & 75.00 & 75.00 \\
  \hline
   Batch 12			& 53.33 & \cellcolor{green!25} 80.70 & 75.52 & 67.76 \\
  \hline 
  \textbf{Overall} 	& \textbf{82.31} &  \cellcolor{olive!25} \textbf{85.18} & \textbf{85.00} & \textbf{81.02} \\
  \hline
\end{tabular}
}
\end{table*}

\section{\uppercase{Discussion}}
\label{sec:discussion}

\begin{figure*}
  \centering
   \begin{subfigure}[b]{0.20\textwidth}
         \centering
         \includegraphics[width=\textwidth]{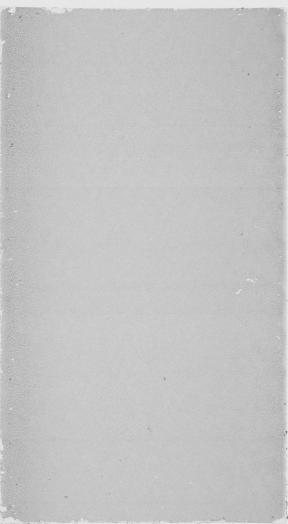}
         \caption{}
         \label{fig:false-labeling-example-a}
     \end{subfigure}
     \hfill
     \begin{subfigure}[b]{0.20\textwidth}
         \centering
         \includegraphics[width=\textwidth]{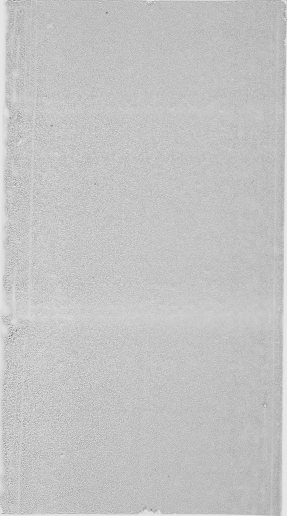}
         \caption{}
         \label{fig:false-labeling-example-b}
     \end{subfigure}
     \hfill
     \begin{subfigure}[b]{0.20\textwidth}
         \centering
         \includegraphics[width=\textwidth]{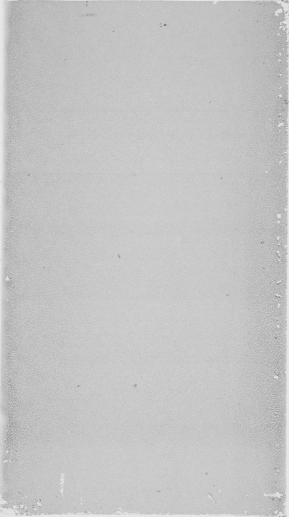}
         \caption{}
         \label{fig:false-labeling-example-c}
     \end{subfigure}
     \hfill
     \begin{subfigure}[b]{0.20\textwidth}
         \centering
         \includegraphics[width=\textwidth]{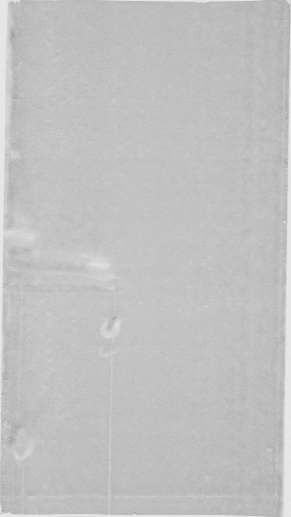}
         \caption{}
         \label{fig:false-labeling-example-d}
     \end{subfigure}
  \caption{Labeling fuel cell electrodes is a non-trivial task which can be prone to subjectivity. Figure \ref{fig:false-labeling-example-a} and \ref{fig:false-labeling-example-b} are examples of false positives and figure \ref{fig:false-labeling-example-c} and \ref{fig:false-labeling-example-d} are examples of false negatives found during evaluation of conversion method 2. As can be seen the difference between e.g. figure \ref{fig:false-labeling-example-a} and \ref{fig:false-labeling-example-c} is difficult to see and the mis-classification might as well be due to faulty annotations.}
  \label{fig:false-labeling-examples}
 \end{figure*}

Labeling anomalies in X-ray images of fuel cell electrodes is a difficult and time consuming task, which requires expertise in the specific domain and knowledge about the severity and consequences different anomaly types impose to the conductivity of the fuel cell systems in which the electrodes will be used. 
To illustrate the difficulty, a few examples of false positives and false negatives found during the evaluation, can be seen in figure \ref{fig:false-labeling-examples}, which might be cause by faulty annotations.

Generally, when labeling samples for a binary classification problem, the person labeling has to make a decision on which class the sample belong to, for this work, whether an electrode is normal or abnormal.

We find that making this decision, for some samples, is a non-trivial task prone to subjectivity.
While one expert might label the sample as normal another expert might label the same sample as abnormal. 
One possible solution, which was chosen for this work, is to let the more experienced expert make the final decision.

A second solution, could be to distribute the labeling task across a large number of experts or non-experts and let the label with most votes represent the sample. The method has been described and evaluated by \cite{vote-labeling} and it has the potential to remove subjectivity from the labels.
The drawback to this solution is that it can be expensive and time consuming and in some cases a sample might end up having equally many votes for each class in which case an additional solution for these cases needs to be found. 

Other solutions could be to simply exclude such samples from the dataset or to introduce a third class which represents samples which are undecidable, if one or more experts disagree.
Such a class would in our case be very small and cause a highly imbalanced dataset.

%
%

\section{\uppercase{Conclusions}}
\label{sec:conclusion}
This paper proposed an anomaly detector using a Deep Convolutional Neural Network, an extended ResNet-34 model, for detecting anomalies in X-ray images of fuel cell electrodes. 
For this purpose a dataset with normal and abnormal fuel cell electrodes X-ray images was created.
The dataset consists of 12 batches of images with a total of 714 X-ray images. 
The anomaly detector is used by the company Serenergy for automatizing a time-consuming manual quality control of the fuel cell electrode X-ray images.
The anomaly detector was trained and evaluated through cross-validation where a single batch of images is used as test set and the remaining batches are used as training set.
The proposed anomaly detector was trained and evaluated using $12\times50$ epochs with a Nvidia GeForce 1080 GTX GPU and the PyTorch deep learning framework.
We compared 16-bit to 8-bit conversion methods for pre-processing the X-ray images.
We find that performing histogram equalization with upper- and lower bounds set by the maximum and minimum pixel values calculated across the entire dataset achieve a better performance than when using local maximum and minimum as upper- and lower bounds calculated for each individual image. 
We achieve a balanced accuracy of 85.18\%. 

In the future, we will continue to explorer approaches for performing more accurate anomaly detection in X-ray images. Potential improvements could be achieved by using variations of weighted cross-entropy loss and data augmentation to cope with the imbalanced dataset and by utilizing different histogram equalization methods e.g. the DHE algorithm. 
Further, we see a great potential in using CNNs pre-trained on large-scale gray-scale image datasets for classifying X-ray images. Whereas most CNNs today are pre-trained on RGB image datasets e.g. ImageNet. 
  
\vfill
\section*{\uppercase{Acknowledgements}}
We would like to thank Serenergy for their contributions, collaboration and dataset for this paper. We would also like to thank Ambolt Aps for initiating and facilitating the collaboration with Serenergy.

\bibliographystyle{apalike}
{\small
\bibliography{report-unanonymous}}



\end{document}